\documentclass[]{interact}

\usepackage[hyphens]{url}  % DO NOT CHANGE THIS
\usepackage{algorithm}
\usepackage{algorithmic}

\usepackage{newfloat}
\usepackage{listings}

\usepackage{epstopdf}% To incorporate .eps illustrations using PDFLaTeX, etc.
\usepackage{subfigure}% Support for small, `sub' figures and tables

\usepackage{natbib}% Citation support using natbib.sty
\bibpunct[, ]{(}{)}{;}{a}{}{,}% Citation support using natbib.sty
\renewcommand\bibfont{\fontsize{10}{12}\selectfont}% Bibliography support using natbib.sty

\theoremstyle{plain}% Theorem-like structures provided by amsthm.sty

\theoremstyle{definition}

\theoremstyle{remark}

\begin{document}

\articletype{ARTICLE TEMPLATE}

\title{Controlling Travel Path of Original Cobra}

\author{
\name{Mriganka Basu RoyChowdhury\textsuperscript{a} and Arabin K Dey\textsuperscript{b}\thanks{CONTACT Arabin K Dey. Email: arabin.k.dey@gmail.com}}
\affil{\textsuperscript{a} Department of Mathematics, University of California, Berkley; \textsuperscript{b} Department of Mathematics, IIT Guwahati;}
}

\maketitle

\begin{abstract}
In this paper we propose a kernel based COBRA which is a direct approximation of the original COBRA.  We propose a novel tuning procedure for original COBRA parameters based on this kernel approximation.   We show that our proposed algorithm provides much better accuracy than other COBRAs and faster than usual Gridsearch COBRA.  We use two datasets to illustrate our proposed methodology over existing COBRAs.
\end{abstract}

\begin{keywords}
 Ridge Regression; LASSO; Decision Tree; COBRA; Gradient Descent; gridSearchCV; Randomized SearchCV.
\end{keywords}

\section{Introduction}
\label{sec1}

 Combined regression strategy (COBRA) [\cite{biau2013cobra}, \cite{biau2016cobra}] is a popular ensemble technique to improve accuracy of the prediction in a regression problem.  The appeal of original COBRA reduces due to its discrete structure in weight calculation, which increases the computational burden as it adapts a Grid search approach to choose its optimal parameters.  A kernel-based ensemble learning [\cite{guedj2020kernel}] and its implementation through python [\cite{guedj2017pycobra}] makes faster implementation of COBRA.  However, it uses a different set of parameters independent of the threshold parameters of the original COBRA estimator that consists of a smooth weight function.  In our paper, we plan to tame the original COBRA by choosing a suitable kernel that depends on the threshold parameter of the original COBRA.  The methodology will help us to train faster than Grid search method for choosing its optimal parameters.  In fact the methodology can help us to achieve much better the accuracy level than similar algorithms that take different weak learners like Ridge regression, Lasso, and Decision Tree inside the strategy.  

 There is a wide variety of ensemble algorithms [\cite{dietterich2000ensemble}, \cite{giraud2014introduction}, \cite{shalev2014understanding}], with a crushing majority devoted to linear or convex combinations.   A non-linear way of combining estimators is available by [\cite{mojirsheibani1999combining}].   Ensemble algorithms (e.g. Adaboost [\cite{solomatine2004adaboost}]) also use threshold parameters.  However, Adaboost uses an exponential error function, whereas COBRA builds its territory under the square error loss function.  Recently, \cite{fan1999asymptotic} also developed asymptotic distribution of combining regression estimator.  

 We organize the paper in the following way.  In section 2, we provide the conventional structure of COBRA.  Our proposed strategy of tuning the original COBRA parameter is available in section 3.  Section 4 describes all datasets.  The detailed illustration of real-life implementation is avaiable in section 5. We conclude the paper in section 6.
 
\section[]{Usual CoBRA}

  The combined regression strategy (COBRA) combines predictions from different weak learners to make the prediction.  
Suppose we have a data set of the following form : $(X_{i}, y_{i})\in R^{d + 1}; i = 1, \ldots, N$.  We divide the data set into training and test set, where test set observations are, $(X_{i}, y_{i})\in R^{d + 1}; i = 1, \ldots, l$.  Let's denote the M weak learners that COBRA uses are, say $r_{1}(\cdot), r_{2}(\cdot), r_{3}(\cdot), \ldots, r_{M}(\cdot) : R^{d} \rightarrow R.$  COBRA comes with an objective to find an ensembled strategy to combine all the regressors to predict a new observation with better accuracy.  
Mathematically, the COBRA strategy expresses a predictive estimator $T_{n}$ in the following way :

$$T_{n}(r_{k}(\textbf{x})) = \sum_{i = 1}^{l} W_{n, i}(\textbf{x})Y_{i}, ~~~~~~ x \in R^{d}$$  

where the random weights $W_{n, i}(\textbf{x})$ take the form :
$$ W_{n, i}(\textbf{x}) = \frac{1_{\cap_{m = 1}^{M} {|r_{k, m}(x) - r_{k, m}(X_{i})| \leq \epsilon_{l}}}}{\sum_{j = 1}^{l} 1_{\cap_{m =1}^{M} {|r_{k, m}(x) - r_{k, m}(X_{i})| \leq \epsilon_{l}} }}.$$

 In the above expressions, it uses its predictor at particular instances ($x$) as weighted sum of those test sample $Y_{i}$s where weights are calculated based on number of test sample instances which are in the neighbourhood of predictions of made by each predictors at the particular instances. 

  Here $\epsilon_{l}$ is a positive number and $\frac{0}{0} = 0$ (by convention).  We can provide an alternative way to express COBRA when all original observations are invited to have the same, equally valued opinion on the importances of the observation $X_{i}$ (within the range of $\epsilon_{l}$) for the corresponding $Y_{i}$.  This unanimity constraint may be relaxed by imposing, for example, a fixed fraction $\alpha \in \{ \frac{1}{M}, \frac{2}{M}, \cdots, 1 \}.$ of the machines agrees on the importance of $X_{i}$.  In that case, the weights take the more sophisticated form :
$$ W_{n, i}(x) = \frac{1_{\sum_{m = 1}^{M} 1_{|r_{k, m}(x) - r_{k, m}(X_{j})| \leq \epsilon_{l} } \geq M\alpha}}{\sum_{j = 1}^{l} 1_{\sum_{m = 1}^{M} 1_{|r_{k, m}(x) - r_{k, m}(X_{j})| \leq \epsilon_{l} } \geq M\alpha}}  $$

\section{Our Proposed Strategy : Faster training than GridSearch in COBRA}

  The literature already contains research that discusses COBRA for its faster move than Grid Search in the following way :
\begin{enumerate}
\item  Approximate the indicator function with some suitable smooth kernel.
\item  Exploit the smoothness to build a gradient decent type algorithm.
\end{enumerate}
There are issues in the current research taking this approach.   They use exponential moving average type approximation, which is already available in the original COBRA paper.  Guedj and Desikan exploits this expression and replace the exponent with euclidean distance between $r^{(k)}_{m}(X^{(l)}_{i})$ and $r^{(k)}_{m}(X)$. Therefore this approach is separate kernel-based ensembled learning.

 We look at the solution from a slightly different angle.  We stick to the original COBRA with its discrete nature but use kernel only to tune the hyper-parameter $\epsilon$.  The earlier kernel-based approach does not directly connect this $\epsilon$ in a discrete setup and its hyperparameter.  Therefore in our smooth kernel, we propose to use the following specific approximation.
\begin{enumerate}
\item A direct smooth approximation of the indicator function based on the variable \\
$\max_{m = 1, \cdots, M} |r^{(k)}_{m}(X^{(l)}_{i}) - r^{(k)}_{m}(X)|$.

\item A two stage approximation which involve $L_1$ norm between $r^{(k)}_{m}(X^{(l)}_{i})$ and $r^{(k)}_{m}(X)$ instead of euclidean distance between $r^{(k)}_{m}(X^{(l)}_{i})$ and $r^{(k)}_{m}(X)$ proposed by Guedj and Desikan.
\end{enumerate}    

 Since our target is to construct a smooth function of $\epsilon$, the above mentioned approximations are more effective than usual exponential moving average, which is nothing but a special case of softmax regression type expression.  As our aim is to work with original COBRA and simultaneously making a faster tuning of $\epsilon$, we adapt this novel kernel COBRA (con-COBRA) and compare all the results in different data set. 
 
\subsection{Proposed Algorithm}

 Note that $W$ can also be written as : $$ W_{i}(X) = 1_\{\{\max_{m = 1, \cdots, M} |r^{(k)}_{m}(X^{(l)}_{i}) - r^{(k)}_{m}(X)| < \epsilon \}\} $$
 Now consider the following softening of $1_\{\max_\{ x_{1}, x_{2}, \cdots, x_{m} \} < \epsilon \}$  for some given large values of steepness parameter $\beta$ which can able to approximate the indicator function well,
$$ \phi_{\beta}(x_{1}, x_{2}, \cdots, x_{m}; \epsilon) = \frac{\exp(\beta \epsilon)}{\exp(\beta \epsilon) + \sum_{j = 1}^{m} \exp(\beta x_{j})} $$  

Using this, we approximate $W$ as follows :
$ W_{i}^{\beta}(X) = \phi_{\beta}(|r^{(k)}_{1}(X^{(l)}_{i}) - r^{(k)}_{1}(X)|, |r^{(k)}_{2}(X^{(l)}_{i}) - r^{(k)}_{2}(X)|, \cdots, |r^{(k)}_{M}(X^{(l)}_{i}) - r^{(k)}_{M}(X)|; \epsilon) $

Our goal is to find $\epsilon$ by optimizing the squared loss over a training dataset of size n, ${(\tilde{X}_{j}, \tilde{Y}_{j})}_{j = 1}^{n}$ :
$$ SL(\epsilon) = \sum_{j = 1}^{n} (g(r^{(k)}(\tilde{X}_{j})) - \tilde{Y}_{j})^{2}$$
which is equivalent to optimizing the mean-squared error.  To this end, we use gradient descent.  We compute the derivative of $\frac{dSL}{d\epsilon}$ as follows : 
Define $$ w_{ji}(\epsilon) = \frac{W^{\beta}_{i}(\tilde{X}_{j})}{\exp(\beta \epsilon)}.$$  Consequently the prediction for j-th data point is :
$$ p(\epsilon; \tilde{X}_{j}) = g(r^{(k)}(\tilde{X}_{j})) = \frac{\sum_{i = 1}^{l} w_{ji}(\epsilon) Y^{(l)}_{i}}{\sum_{i = 1}^{l} w_{ji}(\epsilon)}$$

The squared error loss becomes $$ SL(\epsilon) = \sum_{j = 1}^{n} (p(\epsilon; \tilde{X}_{j}) - \tilde{Y}_{j})^{2} $$
Differentiating with respect to $\epsilon$, we can obtain :
$ SL^{'}(\epsilon) = \sum_{j = 1}^{n} (p(\epsilon; \tilde{X}_{j}) - \tilde{Y}_{j})p^{'}(\epsilon; \tilde{X}_{j})$

where the derivative $p^{'}(\epsilon; \tilde{X}_{j})$ is with respect to $\epsilon$.  We now compute,
\begin{eqnarray}
&& p^{'}(\epsilon; \tilde{X}_{j}) \nonumber\\ & = & \frac{(\sum_{i = 1}^{l}w^{'}_{ji}(\epsilon) Y^{(l)}_{i})A - (\sum_{i = 1}^{l}w_{ji}(\epsilon) Y^{(l)}_{i}) B}{(\sum_{i = 1}^{l} w_{ji}(\epsilon))^{2}} \nonumber
\end{eqnarray}

where, $A = \sum_{i = 1}^{l} w_{ji}(\epsilon)$, $ B = \sum_{i = 1}^{l} w^{'}_{ji}(\epsilon).$

But $w^{'}_{ji}(\epsilon) = -\frac{\beta \exp(\beta\epsilon)}{w^{2}_{ji}(\epsilon)}$. Substituting this in the above, we get :
$ p^{'}(\epsilon; X_{j}) = [-\beta\exp(\beta\epsilon)]\frac{(\sum_{i = 1}^{l}w^{2}_{ji}(\epsilon) Y^{(l)}_{i}) A - (\sum_{i = 1}^{l}w_{ji}(\epsilon) Y^{(l)}_{i}) D}{(\sum_{i = 1}^{l} w_{ji}(\epsilon))^{2}} $
$ D = \sum_{i = 1}^{l} w^{2}_{ji}(\epsilon) $	
Using these we can compute $SL^{'}(\epsilon)$.  After that, we can use gradient descent to find out optimal $\epsilon$.

\subsection{Alternative Similar Variation}

Using this, we approximate a slightly different $W$ as follows :
$ W_{i}^{\beta}(X) = \phi_{\beta}(|r^{(k)}_{1}(X^{(l)}_{i}) - r^{(k)}_{1}(X)|, |r^{(k)}_{2}(X^{(l)}_{i}) - r^{(k)}_{2}(X)|, \cdots, |r^{(k)}_{M}(X^{(l)}_{i}) - r^{(k)}_{M}(X)|; \epsilon) $
where
$$ \phi_{\beta}(x_{1}, x_{2}, \cdots, x_{M}; \epsilon) = \frac{e^{\beta \epsilon}}{e^{\beta \epsilon} + e^{\beta \max_{m = 1, \cdot, M } x_{j}}} $$  

  In this case, derivative of $\frac{d SL}{d\epsilon}$ will take exactly the same form as that of proposed algorithm except the fact that $W_{i}^{\beta}(X)$ will take the above form.  
    
\subsection{Some comments and Comparison of the proposed structure with existing other Kernel structure : }   
  
  Usage of kernel in COBRA structure is not new.  Considering a kernel function as weight is discussed separately in some papers.  Our proposed structure is similar to the kernel structure and can question readers with the novelty of the concepts.  Some keypoints in this structure is this Kernel proposition consists of parameter values which has connection with the original threshold version of COBRA.  Thus, we can connect discrete version with its continuous in terms of the same parameters.  We exploit this conversion or structural property where we propose to tune the original version of the COBRA parameters but with the help of converted kernel proposed in this case.   In a different note, other kernel approaches consider parameters where parameters are implicit/not directly visible function of original COBRA parameters.    

\section{DataSet Description} 

  we consider two popular datasets on Housing prices (Boston Housing Price Dataset and California Housing Price Dataset) publicly available in latest version of python module scikit-learn.  Someone can also download the dataset from multiple publicly available other sources/archive (e.g. \url{http://lib.stat.cmu.edu/datasets/}, \url{https://www.dcc.fc.up.pt/~ltorgo/Regression/} etc. 

\subsection{Boston Housing Price Dataset}  

  This dataset contains information collected by the U.S Census Service concerning housing in the area of Boston Mass.  A quick summary of the data set is follows : Samples total := 506; Dimensionality := 13; Features := real, positive [\cite{harrison1978hedonic}].  

\subsection{California Housing Price Dataset}  

  In this dataset the target variable is the median house value for California districts, expressed in hundreds of thousands of dollars (\$100,000).

  This dataset was derived from the 1990 U.S. census, using one row per census block group.  In this sample a block group on average includes 1425.5 individuals living in a geographically compact area. Naturally, the geographical area included varies inversely with the population density. We computed distances among the centroids of each block group as measured in latitude and longitude. We excluded all the block groups reporting zero entries for the independent and dependent variables.  The final data contained 20,640 observations on 9 characteristics (8 numeric, predictive attributes and the target).

  Note that an household is a group of people residing within a home. Since the average number of rooms and bedrooms in this dataset are provided per household, these columns may take surprisingly large values for block groups with few households and many empty houses, such as vacation resorts.
  
\section{Numerical Results} 

 Since both the proposed algorithm and its other variation are almost the same, we choose to opt for one of them for our implementation.  We take 3 weak learners in this experiment.  They are Ridge Regression, LASSO, and Decision Tree.  We take mean square error and $R^2$ as two benchmarks to compare the proposed controlled COBRA with COBRA in a gridsearch and randomized search approach.  Optimized parameter selection for this implementation uses 5 fold cross-validation.  All calculations consider 80\% train sample and 20 \% test sample. 
 
 All codes are run on the computers with the following configurations : (i) Intel(R) Core(TM) i5-6200U CPU 2.30 GHz.  (ii) Ubuntu 20.04 LTS OS, and (iii) 12 GB Memory.  All codes in written in python 3.9 in Jupyter Notebook.  All codes can be made available on request to authors.  
 
\subsection{Boston Housing Price data}     
\begin{figure}[ht!]
\begin{center}
    \includegraphics[width = 0.9\columnwidth]{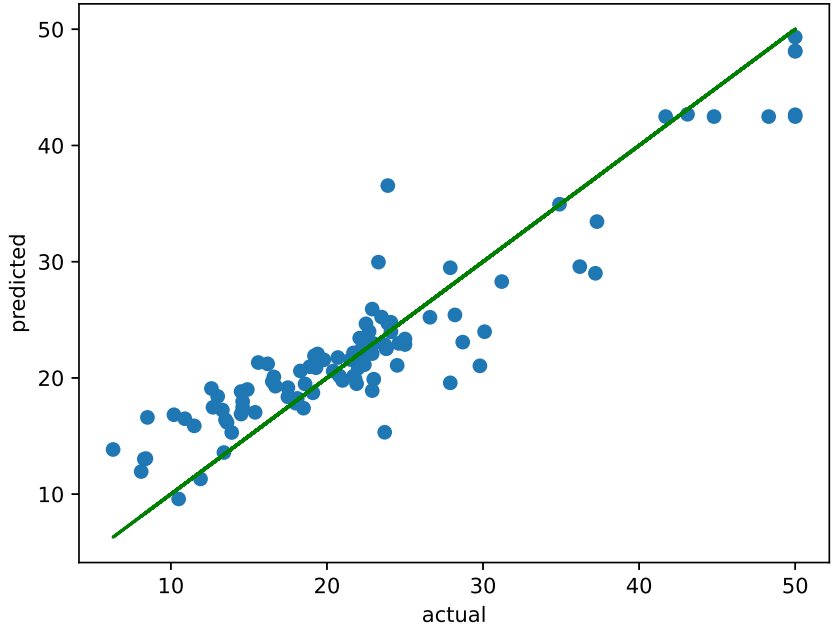}
\caption{Integration of Boston Housing Price through Newly proposed controlled CoBRA \label{Fig.1}}
\end{center}
\end{figure}

Table-\ref{table:bostonControlledCOB}, Table-\ref{table:bostonGridCOB} and Table-\ref{table:bostonRSCOB} show detailed comparison and performance for each weak learners and COBRA.  We can observe the drastic improvement in MSE and $R^2$ for each COBRA implementation over its weak learners as expected.   Figure-\ref{Fig.1} shows the scatter plot of the original Boston housing price data and its predicted values.  The graph proposes the best-fitted curve possible in this context.  We can obtain the optimal threshold parameter for controlled COBRA as 4.838 (approx). 

\begin{table}[ht!]
\begin{tabular}{ccc}  \hline
Estimator \#0: & MSE = 25.201469420 & R2 = 0.744995091 \\ \hline
Estimator \#1: & MSE = 31.879621721 & R2 = 0.677421190  \\ \hline
Estimator \#2: & MSE = 19.555490196 & R2 = 0.802124793  \\ \hline
COBRA & MSE = 12.686645798 & R2 = 0.871628241  \\ \hline
\end{tabular}
\caption{Table showing Benchmarks in different weak leaners and COBRA through proposed controlled COBRA for Boston Housing Price data}
\label{table:bostonControlledCOB}
\end{table} 

\begin{table}[ht!]
\begin{tabular}{ccc}  \hline
Estimator \#0: & MSE = 25.201469420 & R2 = 0.744995091  \\ \hline
Estimator \#1: & MSE = 31.879621721 & R2 = 0.677421190  \\ \hline
Estimator \#2: & MSE = 19.555490196 & R2 = 0.802124793  \\ \hline
Estimator \#3: & MSE = 71.758505882 & R2 = 0.273900623  \\ \hline
COBRA & MSE = 16.475750277 & R2 = 0.833287610  \\ \hline
\end{tabular}
\caption{Table showing Benchmarks in different weak leaners and COBRA through GridsearchCV COBRA for Boston Housing Price data}
\label{table:bostonGridCOB}
\end{table} 

\begin{table}[ht!]
\begin{tabular}{ccc}  \hline
Estimator \#0: & MSE = 25.201469420 & R2 = 0.744995091  \\ \hline
Estimator \#1: & MSE = 31.879621721 & R2 = 0.677421190  \\ \hline
Estimator \#2: & MSE = 19.555490196 & R2 = 0.802124793  \\ \hline
COBRA & MSE = 14.601340698 & R2 = 0.852254109  \\ \hline
\end{tabular}
\caption{Table showing Benchmarks in different weak leaners and COBRA through Randomized SearchCV COBRA for Boston Housing Price data}
\label{table:bostonRSCOB}
\end{table} 

\subsection{California Housing Price data}  
\begin{figure}[ht!]
\begin{center}
    \includegraphics[width = 0.9\columnwidth]{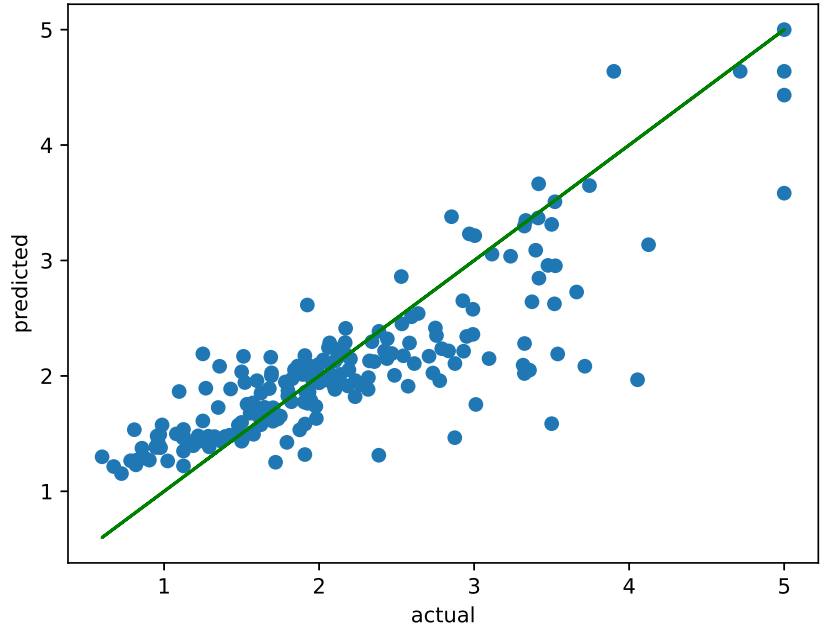}
\caption{Integration of California Housing Price through Newly proposed controlled CoBRA \label{Fig.2}}
\end{center}
\end{figure}

We carry the data analysis based on 1000 observations only.  We do not consider all observations to obtain the implementation.   We can also choose to consider two parameters instead of just one parameter.  Figure-\ref{Fig.2} shows the scatter plot of the original data and predicted values.  The final prediction graph proposes the best-fitted curve possible in this context.  The Table-\ref{table:caliControlledCOB}, Table-\ref{table:caliGridCOB}, Table-\ref{table:caliRSCOB} show performance of all weak learners and COBRA with respect to two benchmarks.
  
\begin{table}[ht!]
\begin{tabular}{ccc}   \hline
Estimator \#0: & MSE = 0.283177 & $R^2$ = 0.654688  \\ \hline
Estimator \#1: & MSE = 0.341271 & $R^2$ = 0.583848  \\ \hline
Estimator \#2: & MSE = 0.289597 & $R^2$ = 0.646860  \\ \hline
COBRA & MSE = 0.243286 & $R^2$ = 0.703332  \\ \hline
\end{tabular}
\caption{Table showing Benchmarks in different weak leaners and COBRA through proposed controlled COBRA for California Housing Price data}
\label{table:caliControlledCOB}
\end{table} 

\begin{table}[ht!]
\begin{tabular}{ccc}  \hline
Estimator \#0: & MSE = 0.283177 & $R^2$ = 0.654688 \\ \hline
Estimator \#1: & MSE = 0.341271 & $R^2$ = 0.583848 \\ \hline 
Estimator \#2: & MSE = 0.289597 & $R^2$ = 0.646860 \\ \hline 
COBRA & MSE = 0.242612 & $R^2$ = 0.704154  \\ \hline
\end{tabular}
\caption{Table showing Benchmarks in different weak leaners and COBRA through Grid-SearchCV for California Housing Price data}
\label{table:caliGridCOB}
\end{table} 

\begin{table}[ht!]
\begin{tabular}{ccc}  \hline
Estimator \#0: & MSE = 0.283177 & R2 = 0.654688  \\ \hline
Estimator \#1: & MSE = 0.341271 & R2 = 0.583848  \\ \hline
Estimator \#2: & MSE = 0.289597 & R2 = 0.646860  \\ \hline
COBRA & MSE = 0.242519 & R2 = 0.704267  \\ \hline
\end{tabular}
\caption{Table showing Benchmarks in different weak leaners and COBRA through Randomized-SearchCV for California Housing Price data}
\label{table:caliRSCOB}
\end{table} 
 
\begin{table}[ht!]
\begin{tiny}
 \begin{tabular}{cccc}
  Runtime for   &  Runtime for  & Runtime for usual & Runtime for   \\ \hline 
 different algorithms & usual COBRA  & COBRA  & controlled  \\ \hline
                      &   (Gridsearch)                &  (Randomized search)        &   COBRA       \\ \hline
  Data Set Chosen     &                   &          &          \\ \hline
  Boston Housing      &      28.623s   &   6.805s & 15.184s   \\
  Price data &                   &          &                        \\ \hline
  California Housing      &   89.426s    &   19.1116s   &   32.506s  \\ 
  Price data  &                   &          &                        \\ \hline
 \end{tabular}
 \caption{Table showing Runtime for different COBRAs on two different datasets}
\label{table:runtime}
\end{tiny}
\end{table}

 From Table-\ref{table:runtime} and other tables above, we observe that our proposed controlled COBRA provides the best metrics with an improved run time.  However, the proposed algorithm is not as fast as COBRA with RandomizedCV.  Therefore, someone needs to choose to opt for the proposed COBRA or COBRA with RandomizedCV depending on the priority set by the context of the problem.  According to our current study, if the time is not too heavy price in the context of the problem, the proposed COBRA always provides the regression with the best accuracy. The merit that this approach possesses over RandomizedCV is RandomizedCV may not always capable to reach the best optimal solution. 
 
\section{Conclusion}  The proposed kernel structure is direct extension of the original COBRA.  The connection of the parameters helps the structure to tune original COBRA parameter more systematically.  The benchmark values indicate that the proposed COBRA achieves the best accuracy and faster than GridsearchCV COBRA.  However, it is not as fast as RandomizedCV.  The practitioner needs to take a call where they should keep their priority and accordingly, someone can either choose the proposed COBRA or RandomizedCV CoBRA.  The work can further be extended to estimate a different functional structure or in a sequential estimation problem.

\section*{Disclosure statement}

 There is no conflict of interest for this work with anyone.

\section*{Funding}

 The work is not related to any funding.

\section*{Notes on contributor(s)}

\section*{Data Set and codes}  from multiple publicly available other sources/archive (e.g. \url{http://lib.stat.cmu.edu/datasets/}, \url{https://www.dcc.fc.up.pt/~ltorgo/Regression/}.  We can make all codes available on a separate request to us.

\section{References}

\bibliographystyle{tfcad}
\bibliography{cobra}

\end{document}